\documentclass{article}

\usepackage{arxiv}

\usepackage[utf8]{inputenc} % allow utf-8 input
\usepackage[T1]{fontenc}    % use 8-bit T1 fonts
\usepackage[colorlinks=true, linkcolor=blue, urlcolor=blue, citecolor=blue]{hyperref}       % hyperlinks
\usepackage{url}            % simple URL typesetting
\usepackage{booktabs}       % professional-quality tables
\usepackage{amsfonts}       % blackboard math symbols
\usepackage{nicefrac}       % compact symbols for 1/2, etc.
\usepackage{microtype}      % microtypography
\usepackage{lipsum}		% Can be removed after putting your text content
\usepackage{graphicx}
\usepackage[numbers]{natbib}
\usepackage{doi}

\usepackage{graphicx}
\usepackage{listings}
\usepackage{xcolor}
\usepackage[most]{tcolorbox}
\usepackage{booktabs}  % for nicer lines
\usepackage{multirow}  % optional, if needed
\usepackage{makecell}
\usepackage{caption}
\usepackage{subcaption}  % in your preamble

\tcbset{
    colback=gray!5,
    colframe=black,
    boxrule=0.8pt,
    arc=3pt,
    left=6pt,
    right=6pt,
    top=6pt,
    bottom=6pt
}

\usepackage{xcolor}

\definecolor{keywordcolor}{rgb}{0.0, 0.0, 0.7}
\definecolor{commentcolor}{rgb}{0.0, 0.5, 0.0}
\definecolor{stringcolor}{rgb}{0.58, 0.0, 0.82}

\usepackage{tabularx}   % for tabularx environment
\usepackage{booktabs}   % for \toprule, \midrule, \bottomrule

\title{\emph{AstraAI}: LLMs, Retrieval, and AST-Guided Assistance for HPC Codebases}

\author{
  \href{https://orcid.org/0000-0003-0049-1981}{\includegraphics[scale=0.06]{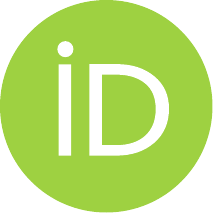}\hspace{1mm}Mahesh Natarajan}\thanks{Corresponding author: MaheshNatarajan@lbl.gov}, \hspace{1mm} \href{https://orcid.org/0000-0002-0747-698X}{\includegraphics[scale=0.06]{orcid.pdf}\hspace{1mm}    
        Xiaoye Li}, \hspace{1mm} and \href{https://orcid.org/0000-0001-8092-1974}{\includegraphics[scale=0.06]{orcid.pdf}\hspace{1mm}    
        Weiqun Zhang} \\
  Applied Mathematics and Computational Research Division \\
  Lawrence Berkeley National Laboratory, Berkeley, CA 94720, USA \\
  \texttt{\{MaheshNatarajan, xsli, WeiqunZhang\}@lbl.gov}
}

\renewcommand{\shorttitle}{\textit{AstraAI}: LLMs, Retrieval, and AST-Guided Assistance for HPC Codebases}

\hypersetup{
pdftitle={\emph{AstraAI}: LLMs, Retrieval, and AST-Guided Assistance for HPC Codebases},
pdfsubject={LLM-assisted code generation for HPC software},
pdfauthor={Mahesh Natarajan, Xiaoye Li, Weiqun Zhang},
pdfkeywords={HPC, large language models, code generation, abstract syntax tree, retrieval augmented generation, AMReX},
}

\begin{document}
\maketitle

\begin{abstract}
We present \emph{AstraAI}, a command-line interface (CLI) coding framework for high-performance computing (HPC) software development. \emph{AstraAI} operates directly within a Linux terminal and integrates large language models (LLMs) with Retrieval-Augmented Generation (RAG) and Abstract Syntax Tree (AST)-based structural analysis to enable context-aware code generation for complex scientific codebases. The central idea is to construct a high-fidelity prompt that is passed to the LLM for inference. This prompt augments the user request with relevant code snippets retrieved from the underlying framework codebase via RAG and structural context extracted from AST analysis, providing the model with precise information about relevant functions, data structures, and overall code organization. The framework is designed to perform scoped modifications to source code while preserving structural consistency with the surrounding code. \emph{AstraAI} supports both locally hosted models from Hugging Face \cite{huggingface} and API-based frontier models accessible via the American Science Cloud \cite{amsc}, enabling flexible deployment across HPC environments. The system generates code that aligns with existing project structures and programming patterns. We demonstrate \emph{AstraAI} on representative HPC code generation tasks within AMReX \cite{AMReX_JOSS,AMReX_IJHPCA1}, a DOE-supported HPC software infrastructure for exascale applications.
\end{abstract}

\section{Introduction}

\noindent High-performance computing (HPC) software plays a critical role in enabling large-scale scientific and engineering simulations. Developing and maintaining HPC code is challenging: large codebases, complex dependencies, low-level parallelism, and strict performance constraints make even minor edits error-prone. Existing AI coding tools provide general suggestions but lack the semantic understanding and correctness guarantees required for safe HPC modifications. This motivates frameworks that combine context-aware retrieval, structured analysis, and controlled code transformations to produce reliable, high-performance scientific software.

High-performance computing (HPC) software development is increasingly leveraging large language models (LLMs) to assist developers in generating, understanding, and optimizing parallel code. Early frameworks such as \textit{chatHPC} \cite{valero2025chathpc,yin2024chathpc} introduced fine-tuned AI agents that provide guidance and code suggestions tailored to parallel systems. Retrieval-Augmented Generation (RAG) approaches further improve code generation by incorporating domain-specific HPC knowledge into the prompting process \cite{gokdemir2025hiperrag,miyashita2024llm}.

Recent systems such as \emph{HPC-GPT} \cite{ding2023hpc} and \emph{HPC-Coder} \cite{nichols2024hpc} demonstrate that fine-tuning LLMs on HPC-specific code and knowledge can significantly improve performance on HPC programming and analysis tasks.

Benchmarking studies provide insight into LLM performance on HPC workloads. Comparisons of models such as GPT-3, Llama-2, and OpenAI Codex highlight differences in code accuracy, parallelization support, and adherence to HPC programming practices \cite{valero2023comparing,godoy2023evaluation}. Broader evaluations examine how language models support HPC software development, assessing capabilities such as correctness, performance, reliability, and trustworthiness across real-world HPC workflows \cite{melone2025llms,godoy2024large,chen2023lm4hpc,kadosh2023scope,chen2024landscape}.

Specialized datasets also play an important role in improving LLM effectiveness; for example, curated examples bridging OpenMP Fortran and \texttt{C++} enable models to learn code equivalence and translation strategies relevant to high-performance applications \cite{lei2023creating}. Overall, integrating AI into HPC workflows raises challenges related to productivity, reliability, and trust, motivating continued research on frameworks, benchmarks, and best practices for safe and effective LLM-assisted HPC development \cite{teranishi2025leveraging,nader2025llm,valero2025chathpc}.

The key novelty of our approach lies in integrating large language models (LLMs) with Retrieval-Augmented Generation (RAG) and Abstract Syntax Tree (AST)-based structural analysis to enable context-aware, controlled code generation for HPC applications. Unlike generic AI-assisted tools, \emph{AstraAI} uses retrieved code examples and AST-derived structural information to guide scoped modifications, ensuring that the generated C\texttt{++} code follows project-specific patterns and maintains structural correctness. This combination supports automated code generation and editing without compromising safety or maintainability, enhancing developer productivity while meeting the strict correctness requirements of HPC workflows.

\section{\emph{AstraAI} architecture}
\begin{figure}[htpb!]
\includegraphics[width=\textwidth]{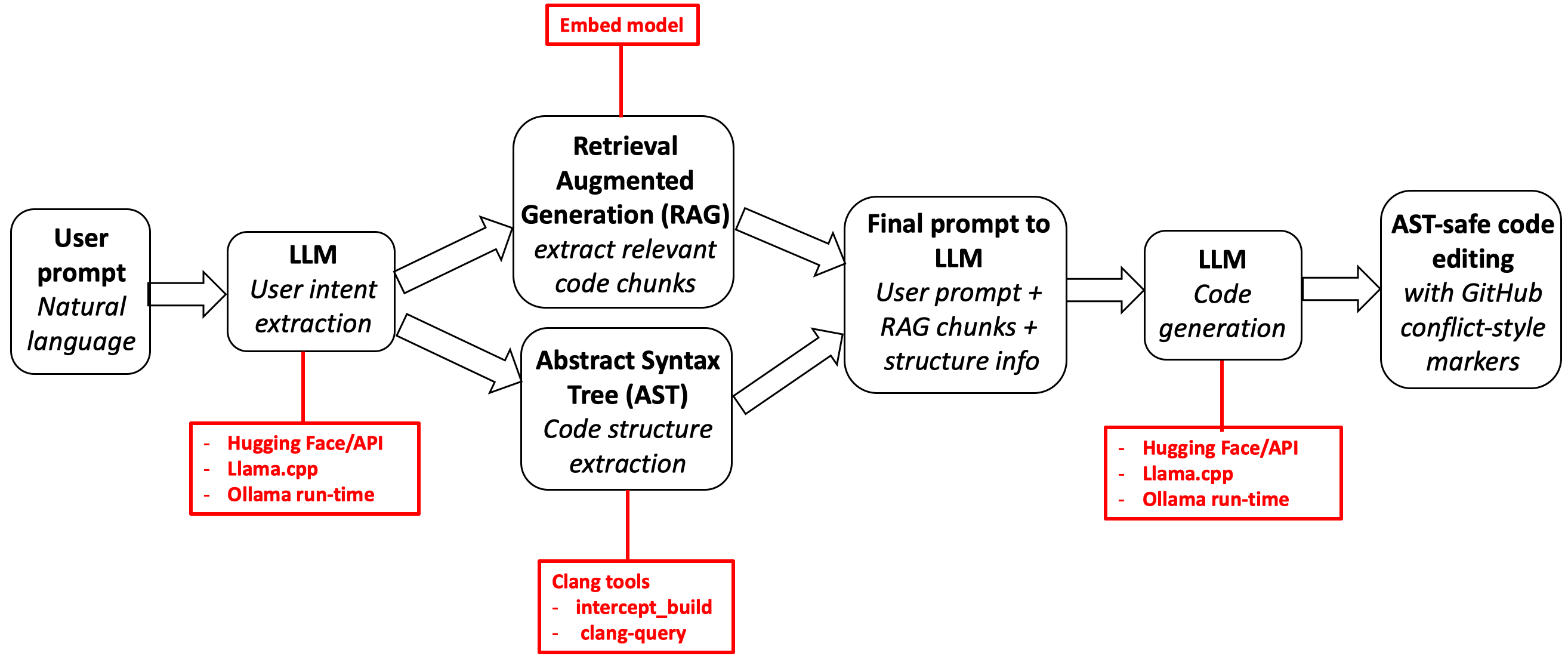}
\caption{End-to-end \emph{AstraAI} workflow: User prompts are interpreted to extract intent, and intent-specific pipelines combine retrieval-augmented context, AST-guided constraints, and controlled code generation or editing. Tools used at each stage are highlighted in red, and the workflow is orchestrated via Python scripts.}
\label{fig:astraai_small}
\end{figure}

%\begin{figure}[htpb]\label{fig:astraai}
%\includegraphics[width=\textwidth]{./Images/\emph{AstraAI}}
%\caption{End-to-end \emph{AstraAI} workflow. User prompts are first classified by intent, after which intent-%specific pipelines combine retrieval-augmented context, AST-guided constraints, and controlled code %generation or editing to ensure safe and reproducible HPC software development.} 
%\end{figure}

Figure~\ref{fig:astraai_small} illustrates the end-to-end workflow for LLM-assisted code generation and editing in HPC applications. The pipeline integrates semantic retrieval and structural program analysis to provide the language model with both contextual and syntactic grounding. The process begins with a user prompt describing the desired functionality or modification. An initial intent extraction stage interprets the user’s goal, followed by two key analysis stages: (i) Retrieval-Augmented Generation (RAG), which retrieves relevant code snippets from a pre-indexed database to provide examples of framework-specific patterns and API usage, and (ii) Abstract Syntax Tree (AST) analysis, which captures the structural organization of the codebase, including classes, member variables, and function signatures. Outputs from the RAG and AST stages are combined with the original prompt to form the final high-fidelity prompt. This composite prompt provides semantic context and syntactic constraints, improving the determinism and correctness of the generated code. The language model then performs code generation, and the output is formatted using AST-safe editing with GitHub conflict-style markers, allowing developers to review and integrate changes while preserving build stability and traceability.
\section{Implementation}

The \emph{AstraAI} framework is organized into two main stages: pre-inference preparation and the inference pipeline. The pre-inference stage consists of one-time or infrequently repeated setup tasks that prepare the model and codebase for efficient downstream processing. These tasks are implemented as standalone Python utilities, each handling a specific preparation step, such as acquiring models from Hugging Face or generating retrieval-ready code chunks for the RAG workflow.

The inference pipeline performs the end-to-end analysis and code generation workflow. It is orchestrated by a central Python driver script that coordinates multiple modular components. The driver sequentially invokes scripts for retrieving relevant code chunks (RAG), performing Abstract Syntax Tree (AST) analysis, constructing prompts, and executing model inference. This modular design improves reproducibility, enables independent testing of pipeline components, and simplifies substitution or extension of individual stages.

\subsection{Pre-inference steps}

\subsubsection{Model access and deployment}
The foundation of the pipeline is the use of LLMs, which can be deployed either locally or accessed via APIs. Local deployment -- using open-source models from ecosystems such as Hugging Face \cite{huggingface} -- provides checkpoints containing network weights, tokenizers, and configuration files, enabling offline operation, reproducibility, and compliance with HPC constraints. For local inference, downloaded models can be converted to formats such as GGUF (GPT-Generated Unified Format) using llama.cpp \cite{llamacpp} and executed through the lightweight runtime engine -- Ollama \cite{ollama}, allowing efficient on-node inference within the \emph{AstraAI} toolchain. Alternatively, \emph{AstraAI} can interact with API-based models, enabling access to frontier-scale LLMs without requiring local installation. In particular, \emph{AstraAI} can connect to models served through the American Science Cloud \cite{amsc}, providing a flexible interface for leveraging large remote models while maintaining the same command-line workflow.

%\paragraph{What is an instruct model?}
%An \textbf{instruction-tuned (instruct) model} is a base pretrained language model that has undergone additional supervised fine-tuning and/or reinforcement learning using datasets formatted as \textit{instruction $\rightarrow$ response} pairs. This alignment process improves task adherence (following user intent), structured reasoning and formatting, conversational reliability, reduced hallucinations in code generation. As a result, instruct models generally outperform base models for interactive coding workflows, debugging assistance, and structured text generation. These models differ in architecture, training corpus composition, and instruction tuning strategy, which leads to variability in language coverage (e.g., C++, Fortran), reasoning depth, and determinism.

%\subsubsection{Conversion to GGUF format}
%To enable efficient local inference, downloaded models are converted to the GGUF (GPT-Generated Unified Format) using the \texttt{llama.cpp} (??) toolchain. GGUF is a binary serialization format designed for lightweight inference engines. It consolidates model weights, tokenizer data, metadata (quantization parameters, architecture details) into a single portable file. Compared with raw Hugging Face checkpoints, GGUF provides faster loading and reduced memory fragmentation and support for multiple quantization levels (e.g., Q4, Q5, Q8).?? This conversion step is critical for achieving practical latency and memory efficiency when running 7B--34B parameter models on shared compute resources.

\subsubsection{Creation of code chunks for RAG}
A central component of the workflow is Retrieval-Augmented Generation (RAG), which injects domain-specific knowledge into the prompt at inference time. To support effective retrieval, a repository of code examples was constructed with fine-grained chunking of source files, metadata annotations covering task type, user-intent descriptions, API/function keywords, and input/output information, and coverage of diverse features and programming idioms. This structure enables semantic search over the codebase, ensuring that retrieved snippets meaningfully complement the user prompt. Efficient retrieval requires transforming both queries and code chunks into vector embeddings using a dedicated embedding model; for this work, the embedding model all-MiniLM was used due to its favorable trade-off between embedding quality and computational cost. Figure~\ref{fig:RAG} illustrates the general structure of the code chunks, including an example of a snippet from the AMReX framework with metadata, serialized into JSON format. In this work, only the \texttt{user\_intent} portion of the metadata (see Fig.~\ref{fig:RAG}) is encoded into embeddings, ensuring that retrieved chunks are semantically aligned with the user’s query rather than other contextual fields.
%In this representation: semantically similar code snippets lie close in vector space, retrieval reduces to nearest-neighbor search, and lexical mismatch (different variable names, formatting) has limited impact. For this work, the embedding model \textit{all-MiniLM} was used due to its favorable trade-off between embedding quality and computational cost. Other commonly used embedding models include: nomic-embed-text, BGE embeddings, jina-embeddings, Qwen embedding models. The embedding model choice directly influences retrieval accuracy, latency, and memory footprint, and therefore constitutes a key design parameter in RAG pipelines.

\begin{figure}[htpb!]
\begin{tcolorbox}[
    colback=red!5,       % background color (5% blue)
    colframe=blue!75!black, % frame color (dark blue)
    boxrule=1pt,          % thickness of frame
    arc=4pt,              % rounded corners
    left=6pt, right=6pt, top=6pt, bottom=6pt,
    title=General structure of code chunk with metadata
]
----- Code -----\\
----- Metadata -----\\
----- Embedding -----
\end{tcolorbox}

\begin{tcolorbox}[
    colback=gray!5,       % background color (5% blue)
    colframe=blue!75!black, % frame color (dark blue)
    boxrule=1pt,          % thickness of frame
    arc=4pt,              % rounded corners
    left=6pt, right=6pt, top=6pt, bottom=6pt,
    title=AMReX code snippet
]
\begin{lstlisting}
// AI_METADATA
// task_type: MULTIFAB_FILL_PARALLELFOR
// user_intent:
// 1) Fill the field data structure for amrex using a 
//    AMReX-style ParallelFor loop
// 2) Fill a multifab using ParallelFor
// 3) Populate the multifab with a parallelfor loop
// 4) For loop in amrex-style GPU-enabled way
// keywords: MFIter, ParallelFor, Array4, GPU, validbox
// inputs: MultiFab mf, dx
// outputs: mf data initialized

for(amrex::MFIter mfi(mf); mfi.isValid(); ++mfi){
    const amrex::Box& bx = mfi.validbox();
    const amrex::Array4<amrex::Real>& mf_array = mf.array(mfi);

    amrex::ParallelFor(bx, [=] AMREX_GPU_DEVICE(int i, int j, int k){

        amrex::Real x = (i+0.5) * dx[0];
        amrex::Real y = (j+0.5) * dx[1];
        amrex::Real z = (k+0.5) * dx[2];

        amrex::Real r_squared = ((x-0.5)*(x-0.5)+(y-0.5)*(y-0.5)+
                                (z-0.5)*(z-0.5))/0.01;

        mf_array(i,j,k) = 1.0 + std::exp(-r_squared);
    });
}
\end{lstlisting}
\end{tcolorbox}

\begin{tcolorbox}[
    colback=gray!5,       % background color (5% blue)
    colframe=blue!75!black, % frame color (dark blue)
    boxrule=1pt,          % thickness of frame
    arc=4pt,              % rounded corners
    left=6pt, right=6pt, top=6pt, bottom=6pt,
    title=AMReX code chunk in \texttt{.json} format
]
\begin{lstlisting}
{
      "text": // the code chunk
      "metadata": {
        "example": "MultiFab",
        "task_type": "MULTIFAB_FILL_PARALLEL",
        "keywords": "MFIter, ParallelFor, Array4, GPU, validbox",
         ....
        "user_intent": ...
      },
      "embedding": [
        -0.05731915682554245,
        0.014803881756961346,
        ....
        ]
}        
\end{lstlisting}
\end{tcolorbox}
\caption{General structure of the chunks for retrieval (top), an example code snippet in AMReX with the metadata (middle), and the code chunk in \texttt{.json} format (bottom).}
\label{fig:RAG}
\end{figure}

\subsection{Inference pipeline}
The inference workflow is fully automated through Python scripting. Users interact with the system minimally by providing a prompt in a text file and executing a single driver script. This script orchestrates retrieval, structural analysis, prompt construction, and model inference, streaming the generated response directly to the terminal. Users can then choose whether to apply the suggested edits to the code. This design ensures reproducibility and allows seamless integration with both batch and interactive HPC workflows.

\subsubsection{Retrieval of relevant code chunks for RAG}
The first stage of inference performs retrieval-augmented context selection. The user prompt is converted into a vector embedding using the same embedding model (all-MiniLM) used during indexing. Cosine similarity is then computed between the query embedding and the database of code chunk embeddings. Top-ranked chunks are selected and appended to the prompt, providing the language model with semantically relevant examples and implementation patterns. This ensures framework-specific grounding and reduces the likelihood of hallucinating APIs or implementation details.

\subsubsection{Structure extraction using AST analysis}
\begin{figure}[h!]
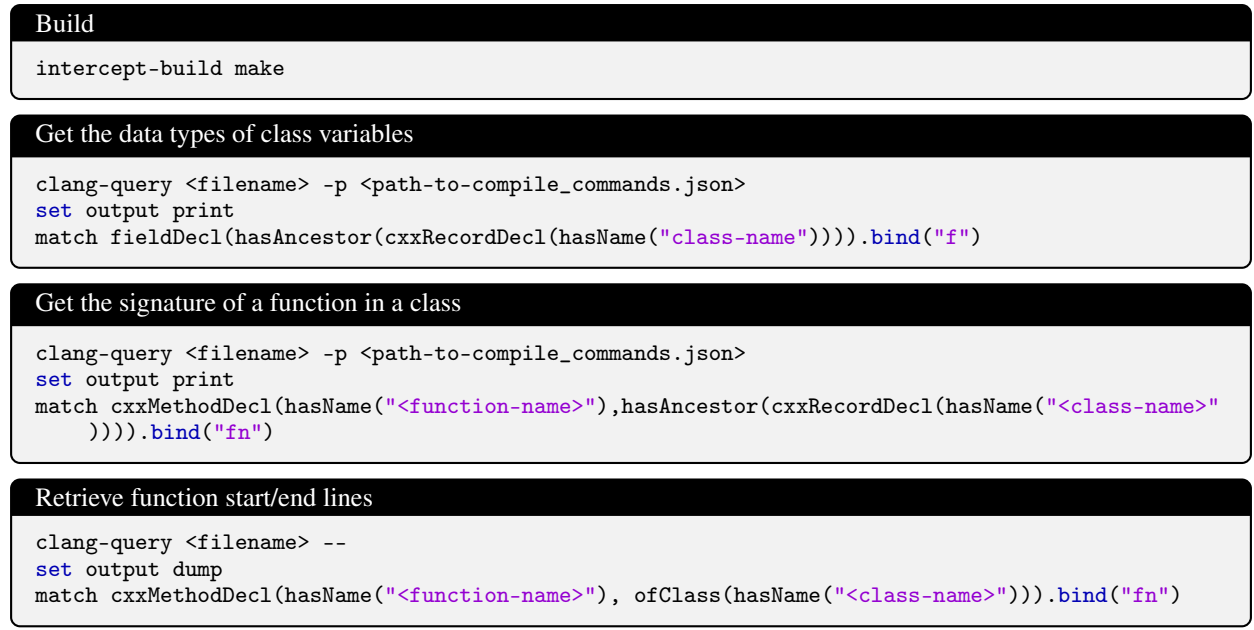

\begin{tcolorbox}[colback=black!5!white,
                  colframe=black!75!black,
                  boxsep=1mm,
                  top=1mm,
                  bottom=1mm,
                  title=Build]
\begin{lstlisting}[language=bash,
                   basicstyle=\ttfamily\small,
                   breaklines=true,
                   aboveskip=0pt,
                   belowskip=0pt]
intercept-build make
\end{lstlisting}
\end{tcolorbox}

\begin{tcolorbox}[colback=black!5!white,
                  colframe=black!75!black,
                  boxsep=1mm,
                  top=1mm,
                  bottom=1mm,
                  title=Get the data types of class variables]
\begin{lstlisting}[language=bash,
                   basicstyle=\ttfamily\small,
                   breaklines=true,
                   aboveskip=0pt,
                   belowskip=0pt]
clang-query <filename> -p <path-to-compile_commands.json>
set output print
match fieldDecl(hasAncestor(cxxRecordDecl(hasName("class-name")))).bind("f")
\end{lstlisting}
\end{tcolorbox}

\begin{tcolorbox}[colback=black!5!white,
                  colframe=black!75!black,
                  boxsep=1mm,
                  top=1mm,
                  bottom=1mm,
                  title=Get the signature of a function in a class]
\begin{lstlisting}[language=bash,
                   basicstyle=\ttfamily\small,
                   breaklines=true,
                   aboveskip=0pt,
                   belowskip=0pt]
clang-query <filename> -p <path-to-compile_commands.json>
set output print
match cxxMethodDecl(hasName("<function-name>"),hasAncestor(cxxRecordDecl(hasName("<class-name>")))).bind("fn")
\end{lstlisting}
\end{tcolorbox}
\begin{tcolorbox}[colback=black!5!white,
                  colframe=black!75!black,
                  boxsep=1mm,
                  top=1mm,
                  bottom=1mm,
                  title=Retrieve function start/end lines]
\begin{lstlisting}[language=bash,
                   basicstyle=\ttfamily\small,
                   breaklines=true,
                   aboveskip=0pt,
                   belowskip=0pt]
clang-query <filename> --
set output dump
match cxxMethodDecl(hasName("<function-name>"), ofClass(hasName("<class-name>"))).bind("fn")
\end{lstlisting}
\end{tcolorbox}
\caption{\texttt{clang-query} command examples to extract structure information of the code.}
\label{fig:clang-query}
\end{figure}

In addition to retrieval, the workflow incorporates static program analysis through Abstract Syntax Tree (AST) extraction. An AST is a tree-structured representation of source code, where each node corresponds to a syntactic construct such as declarations, statements, expressions, or type definitions. Unlike raw text, the AST captures the structural semantics of the program, enabling precise extraction of relationships between program elements. To obtain this representation, the codebase is first compiled using \texttt{intercept-build} (see Fig.~\ref{fig:clang-query}), which generates a \texttt{compile\_commands.json} database containing compiler invocations and flags. Clang tooling reproduces the compilation environment, and the \texttt{clang-query} utility can be used to traverse the AST and extract relevant program information, including class member variables, access specifiers (public, private, protected), function signatures, and function source ranges (start/end lines) (see Fig.~\ref{fig:clang-query}). These structural details are appended to the prompt context, providing the language model with an explicit view of the program interface and reducing ambiguity during code generation or modification tasks. The source range information, in particular, allows precise identification of function locations in the file, enabling automated code modifications and edits. 

\subsubsection{Prompt construction}
\begin{figure}[htpb!]
\centering
\includegraphics[scale=0.8]{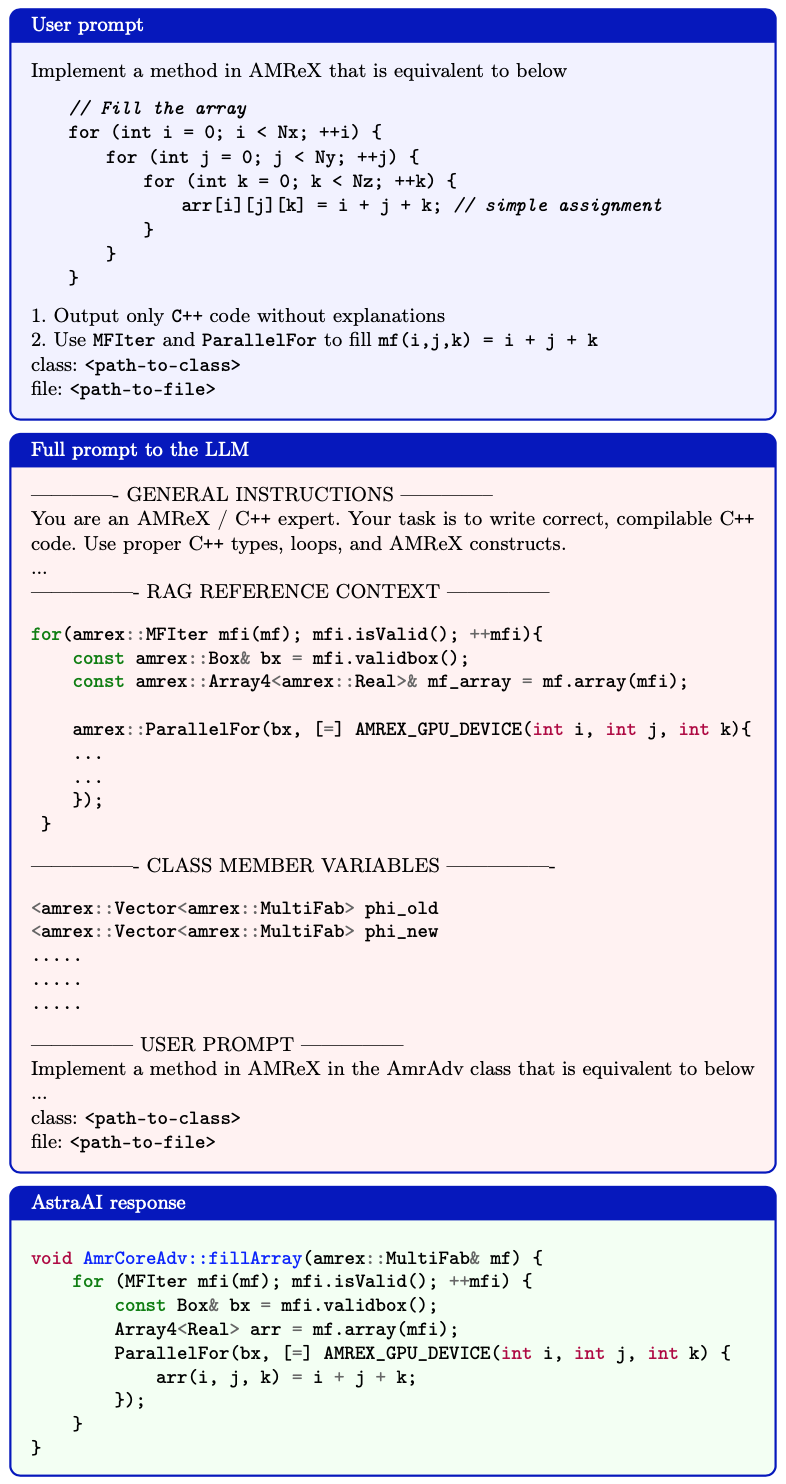}
\caption{An example user prompt that attempts to port a nested \texttt{for} loop in \texttt{C++} into AMReX (top), the full prompt to the LLM (middle), and the result of the inference (bottom).}
\label{fig:astraai_example}
\end{figure}

Following retrieval and AST extraction, the final prompt is assembled by concatenating multiple sources of context. It typically includes general system-level instructions specifying coding style and constraints, retrieved code chunks from the RAG stage, structural information extracted from AST analysis, and the original user query. This layered construction ensures that the LLM simultaneously receives high-level guidance, relevant examples, and precise program structure, improving the determinism and correctness of the generated output. Figure~\ref{fig:full_prompt_structure} illustrates the structure of a full prompt provided to the model. By combining these components -- system instructions, reference snippets, AST-derived structural details, and the user request -- the prompt provides sufficient context to generate code that is both accurate and consistent with the project’s existing structure. Figure~\ref{fig:astraai_example} illustrates a typical user prompt for porting sequential \texttt{C++} code to an AMReX-based framework. The prompt includes a simple array-filling loop in standard \texttt{C++} and specifies that it should be rewritten using \lstinline|MFIter| and \lstinline|ParallelFor| to enable both MPI and GPU parallelism. Additional metadata, such as the relevant class and file location, provides contextual information that guides the language model in generating code consistent with the AMReX programming model.

\subsubsection{Model inference}
The constructed prompt is passed to the language model using the Ollama \cite{ollama} runtime for locally hosted Hugging Face models \cite{huggingface}, or via the American Science Cloud \cite{amsc} API for frontier models. For the local inference, all experiments reported in this work were executed on a single NVIDIA A100 GPU on the Perlmutter system at NERSC (Berkeley Lab). This setup allows \emph{AstraAI} to flexibly leverage either locally hosted models or remote API-based models depending on computational and deployment constraints.

\begin{figure}
\begin{tcolorbox}[
    colback=red!5,       % background color (5% blue)
    colframe=blue!75!black, % frame color (dark blue)
    boxrule=1pt,          % thickness of frame
    arc=4pt,              % rounded corners
    left=6pt, right=6pt, top=6pt, bottom=6pt,
    title=General structure of the high-fidelity prompt
]
----- General instructions -----\\
----- Context derived from Retrieval Augmented Generation (RAG) -----\\
----- Information derived from Abstract Syntax Tree (AST) analysis -----\\
----- User prompt -----
\end{tcolorbox}
\caption{General structure of the high-fidelity prompt.}
\label{fig:full_prompt_structure}
\end{figure}

\section{Evaluation}

We evaluate the performance of \emph{AstraAI} in generating correct and usable code snippets within the AMReX framework, considering several large language models (LLMs), including Hugging Face models \cite{huggingface} -- CodeLlama-13b-Instruct-hf and CodeLlama-34b-Instruct-hf, and frontier models available via the American Science Cloud API \cite{amsc} -- claude-sonnet-4-6-high and gpt-oss-120b. Two evaluation strategies are employed. First, in the baseline evaluation, each model receives only the user prompt and general instructions -- where the prompt requests the implementation of a single function to perform a specific task-- without additional guidance, to measure the effectiveness of the pre-trained model in isolation. Second, in the \emph{AstraAI} evaluation, the user prompt is augmented with RAG and AST guidance to improve the quality and correctness of the generated code.

For quantitative comparison, we compute the cosine similarity between the generated code and benchmark verified code. This metric indicates how closely a model’s output matches the expected solution, providing a measure of improvements achieved by \emph{AstraAI} over the baseline. To reduce the impact of superficial differences in variable naming, local identifiers are normalized in both generated and benchmark code using an AST derived from the tree-sitter C\texttt{++} parser \cite{treesitter}. This procedure identifies variables declared within the function scope while preserving function parameters and framework-specific types, replacing local identifiers with canonical placeholders (e.g., \texttt{VAR1}, \texttt{VAR2}). This ensures that similarity metrics focus on structural and semantic correctness rather than arbitrary naming differences.

The evaluation results are summarized in Table~\ref{tab:baseline} (baseline) and Table~\ref{tab:astraai} (\emph{AstraAI}). Smaller models like CodeLlama-13B and CodeLlama-34B appear to have some pre-training on AMReX or related repositories, but this is insufficient, as reflected by lower baseline cosine similarity scores (Table~\ref{tab:baseline}). With \emph{AstraAI}’s RAG\texttt{+}AST augmentation, these models achieve higher similarity across most benchmarks (Table~\ref{tab:astraai}). The frontier models -- claude-sonnet-4-6-high and gpt-oss-120b -- already show strong baseline performance, indicating substantial prior exposure to AMReX; \emph{AstraAI} improves them in several cases, though not uniformly across all tasks. Variability is expected because multiple valid implementations can satisfy the same specification, so cosine similarity can differ even when outputs are functionally correct.

% ---------- Table 1a ----------
\begin{table}[htpb!]
\centering
\caption{Baseline performance across models showing the cosine similarity metric.}
\label{tab:baseline}
\begin{tabular}{|>{\raggedright\arraybackslash}m{4cm}|>{\centering\arraybackslash}m{2cm}|>{\centering\arraybackslash}m{2cm}|>{\centering\arraybackslash}m{2cm}|>{\centering\arraybackslash}m{2cm}|}
\hline
\multicolumn{1}{|c|}{\makecell[c]{Task}} & \multicolumn{4}{c|}{Baseline (user prompt only)} \\
\cline{2-5}
                             & CodeLlama 13B & CodeLlama 34B &  GPT-OSS-120B & Claude-Sonnet-4.6-High \\
\hline
1. 3D nested \texttt{for} loop to AMReX kernel conversion & 0.32 & 0.99 & 0.92 & 0.94 \\
\hline
2. Creating a copy of a \texttt{MultiFab} from another & 0.60 & 0.65 & 0.82 & 0.87 \\
\hline
3. \texttt{MultiFab} linear combination & 0.77 & 0.83 & 0.90 & 0.85 \\
\hline
4. 7-point finite-difference Laplacian kernel & 0.88 & 0.85 & 0.87 & 0.97 \\
\hline
5. Neumann boundary ghost cell fill & 0.70 & 0.71 & 0.72 & 0.79 \\
\hline
6. Multigrid Poisson solver using AMReX \texttt{MLMG} & 0.47 & 0.82 & 0.81 & 0.89 \\
\hline
\end{tabular}
\end{table}

\vspace{0.5cm}

% ---------- Table 1b ----------
\begin{table}[htpb!]
\centering
\caption{Performance of \emph{AstraAI} across models showing the cosine similarity metric.}
\label{tab:astraai}
\begin{tabular}{|>{\raggedright\arraybackslash}m{4cm}|>{\centering\arraybackslash}m{2cm}|>{\centering\arraybackslash}m{2cm}|>{\centering\arraybackslash}m{2cm}|>{\centering\arraybackslash}m{2cm}|}
\hline
\multicolumn{1}{|c|}{\makecell[c]{Task}} & \multicolumn{4}{c|}{\emph{AstraAI} (user prompt\texttt{+}RAG chunks\texttt{+}AST info)} \\
\cline{2-5}
                             & CodeLlama 13B & CodeLlama 34B &  GPT-OSS-120B & Claude-Sonnet-4.6-High \\
\hline
1. 3D nested \texttt{for} loop to AMReX kernel conversion & 0.78 & 0.92 & 0.91 & 0.92 \\
\hline
2.  Creating a copy of a \texttt{MultiFab} from another & 0.87 & 0.89 & 0.96 & 0.95 \\
\hline
3. \texttt{MultiFab} linear combination & 0.88 & 0.98 & 0.94 & 0.93\\
\hline
4. 7-point finite-difference Laplacian kernel & 0.80 & 1.00 & 0.96 & 0.96 \\
\hline
5. Neumann boundary ghost cell fill & 0.99 & 0.99 & 0.92 & 0.90 \\
\hline 
6. Multigrid Poisson solver using AMReX \texttt{MLMG} & 0.37 & 1.00 & 1.00 & 1.00 \\
\hline
\end{tabular}
\end{table}

\section{Conclusions and future work}

We have presented \emph{AstraAI}, a command-line framework that integrates large language models with Retrieval-Augmented Generation (RAG) and Abstract Syntax Tree (AST) analysis to enable context-aware code generation for HPC software. Our evaluation on AMReX shows that \emph{AstraAI} improves code correctness and structural consistency compared to baseline LLM outputs, particularly for smaller models such as CodeLlama-13B and CodeLlama-34B. By augmenting prompts with repository context and structural information, \emph{AstraAI} enables these models to generate code that closely matches verified benchmark implementations. While large frontier models perform well on AMReX due to its open-source availability, they cannot be expected to generalize reliably to proprietary codebases. \emph{AstraAI} addresses this limitation by enabling context-aware code generation for software without publicly available training data.

For future work, we plan to extend \emph{AstraAI} to support large open-source models using DeepSpeed \cite{deepspeed}, enabling efficient inference through model parallelism. We also plan to add multi-file editing and automated code refactoring while preserving program semantics, allowing \emph{AstraAI} to support larger code modifications. Finally, we will perform more comprehensive evaluations, including compilation success and correctness of numerical results, to better assess \emph{AstraAI}’s ability to generate fully functional scientific code.

The code associated with this work is available at \href{https://github.com/AIForHPC/AstraAI}{https://github.com/AIForHPC/AstraAI}, and will be publicly released after necessary approvals.

\section*{Acknowledgments}
This work was supported in part by the U.S. Department of Energy, Office of Science, Office of Advanced Scientific Computing Research's ModCon under Contract No. AC02-05CH11231. Access to frontier model APIs was facilitated by the American Science Cloud. This research used resources of the National Energy Research Scientific Computing Center (NERSC), a Department of Energy User Facility. We gratefully acknowledge helpful discussions with Bhargav Sriram Siddani and Yingheng Tang (CCSE, LBNL), and thank Akash Vijaykumar Dhruv (Argonne National Laboratory) for his assistance.

\section*{Competing Interests}
The authors have no competing interests to declare that are relevant to the content of this article.

%\bibliographystyle{unsrtnat}
%\bibliography{references}

\end{document}